\documentclass{article}
\usepackage{graphicx}
\usepackage{listings}
\usepackage{bm}

\PassOptionsToPackage{numbers, compress}{natbib}
\usepackage[preprint]{neurips_2026}

\usepackage[utf8]{inputenc}
\usepackage[T1]{fontenc}

\usepackage{amsmath}
\newcommand{\norm}[1]{\left \lVert #1 \right \rVert}

\usepackage{hyperref}
\usepackage{cleveref}
\usepackage{url}
\usepackage{booktabs}
\usepackage{amsfonts}
\usepackage{nicefrac}
\usepackage{microtype}
\usepackage{xcolor}
\usepackage{tikz}
\usetikzlibrary{arrows.meta, calc}
\usepackage{soul}

\title{Aligned Training: A Parameter-Free Method to Improve Feature Quality and Stability of Sparse Autoencoders (SAE)}

\author{
\textbf{Micha{\l}~Brzozowski}$^{1,\dagger}$ \and
\textbf{Neo~Christopher~Chung}$^{1,2}$ \\
[0.6em]
$^{1}$Samsung AI Center, Warsaw, Poland \quad
$^{2}$University of Warsaw, Poland \\
$^{\dagger}$Corresponding author: \texttt{m.brzozowsk3@samsung.com}
}

\begin{document}

\maketitle

\begin{abstract}
Sparse autoencoders (SAEs) are one of the main methods to interpret the inner workings of deep neural networks (DNNs), decomposing activations into higher-dimensional features. However, they exhibit critical shortcomings where a large fraction of features are never activated and are unstable. Despite variants of SAEs that attempt to mitigate these issues, they require additional data, resampling, or training. We propose the \textbf{aligned training}, a parameter-free reparameterization of SAEs that simultaneously improves reconstruction quality, eliminates dead features, and significantly enhances stability across training seeds. Our approach is motivated by an overlooked observation that SAE feature quality, measured by the inner product between encoder and decoder directions (which we call the \textbf{alignment score}), follows a bimodal distribution across all modern architectures. The proposed aligned training enforces a geometric constraint between the encoder and decoder such that their inner product equals one for every feature, which removes a source of degeneracy in the SAE training without adding any hyperparameters. Across multiple models, dictionary sizes, and sparsity levels, the aligned training shows Pareto improvements on the SAEBench benchmarks. Beyond improving dead features, stability and reconstruction, our method readily integrates with techniques in mechanical interpretability such as Top/BatchTop-K architectures and p-Annealing. Overall, the aligned training substantially improves feature quality and stability of SAE without computational complexity or cost.

%We introduce \textbf{aligned training}: a parameter-free reparameterization of sparse autoencoders (SAEs) that simultaneously improves reconstruction quality, eliminates dead features, and significantly enhances stability across training seeds. The method enforces a simple geometric constraint between the encoder and decoder --- that their inner product equals one for every feature --- which removes a source of degeneracy in standard SAE training without adding any hyperparameters.
%Our approach is motivated by an overlooked observation that SAE feature quality, measured by the inner product between encoder and decoder directions (\textit{alignment score}), follows a bimodal distribution across all modern architectures. The alignment score is cheap to compute, requires no additional data or training, and is highly correlated with existing quality proxies. Aligned training eliminates this bimodality by design.
%We demonstrate Pareto improvements over standard and state-of-the-art SAEs on SAEBench benchmarks across multiple models, dictionary sizes, and sparsities, for both ReLU and TopK architectures. Our method reduces dead features to near zero without resampling or auxiliary losses. Furthermore the aligned training produces significantly more stable dictionaries than standard training, where features learnt from independent training runs are more consistent.
\end{abstract}

\section{Introduction}

Deep neural networks (DNN) have made remarkable strides in advancing natural language understanding and generation. Nonetheless, the inner workings of DNNs remain opaque, which limits their development and application in high-risk settings. Mechanistic interpretability has been developed to address this challenge. The superposition hypothesis suggests that a substantially larger number of independent concepts are disentangled in a lower-dimensional activation space of a DNN \citep{elhage2022superposition}. Dictionary learning attempts to recover these concepts by estimating a sparse overcomplete basis.

Sparse autoencoders (SAEs) are among the most popular tools in mechanistic interpretability. Applied to the residual stream of a transformer with sparsity constraints, they decompose activations into interpretable features \citep{huben2023sparse, bricken2023monosemanticity}. Despite the popularity and architectural progress \citep{gao2025scaling, gatedsae2024, rajamanoharan2025jumping}, two problems persist across all variants of SAE: a significant fraction of features are \emph{dead} (never activating on any input) \citep{bricken2023monosemanticity, gao2025scaling, anthropic2024ghostgrads}, and features learned across independent training runs are often \emph{inconsistent} \citep{paulo2026sparse, song2025position}.

The root cause is an overlooked degeneracy. The inner product between each feature's encoder row and decoder column --- which we call the \textbf{alignment score} --- follows a bimodal distribution across all modern SAE architectures (see Appendix~\ref{apx: bimodality}): many features are well-aligned, but a substantial fraction are nearly orthogonal artifacts with alignment scores near zero. Well-aligned features ($a_i \approx 1$) correlate strongly with existing quality proxies such as MCS and autointerpretability, while poorly aligned features correspond to uninterpretable noise. Standard training leaves this degree of freedom entirely unconstrained.

To address this, we introduce \textbf{aligned training}: a parameter-free reparameterization that enforces $W^{enc}_{i,\cdot} \cdot W^{dec}_{\cdot,i} = 1$ for every feature $i$ by construction. This single geometric constraint, motivated by a projection onto an affine hyperplane (Section~\ref{sec: fixing}), simultaneously addresses both failure modes and improves reconstruction quality --- without any auxiliary losses, resampling, or additional hyperparameters.

Through extensive experiments, aligned training is shown to (i) achieve Pareto improvements in reconstruction quality, (ii) reduce dead features to near zero, and (iii) significantly improve feature consistency across seeds --- on SAEBench benchmarks across multiple models, dictionary sizes, sparsity penalties, and activation functions.

%-----------------------------------------------------------------------
\section{Related Work}
%-----------------------------------------------------------------------

\subsection{Sparse Autoencoders} \label{background}

Let $\mathbb{R}^n$ represent the activation space of a deep learning model. The goal of a SAE is to decompose a data point's activation into a sparse linear combination of features from a dictionary of size $m$. Such features have been shown to be more interpretable than the neuron basis \citep{huben2023sparse, bricken2023monosemanticity}.

The encoder and decoder are:
\begin{equation*}
    \mathbf{f}(\mathbf{x}) := \mathrm{ReLU}(W^{enc}\mathbf{x} + \mathbf{b}^{enc}),
    \qquad
    \hat{\mathbf{x}} := W^{dec}\mathbf{f}(\mathbf{x}) + \mathbf{b}^{dec},
\end{equation*}
where $W^{enc} \in \mathbb{R}^{m \times n}$, $W^{dec} \in \mathbb{R}^{n \times m}$. The training loss is
\[
    \mathcal{L} = \underbrace{\norm{\hat{\mathbf{x}} - \mathbf{x}}^2}_{\mathcal{L}_R}
    + \lambda \underbrace{\norm{\mathbf{f}(\mathbf{x})}_1}_{\mathcal{L}_P}.
\]

An important structural property is homogeneity: multiplying the encoder and dividing the decoder by any scalar leaves the reconstruction unchanged while making $\mathcal{L}_P$ arbitrarily small. Two approaches have been proposed to handle this degeneracy.

\paragraph{Normalization} \citep{bricken2023monosemanticity, huben2023sparse} projects decoder columns to unit norm after each gradient step, requiring careful synchronization with the Adam optimizer.

\paragraph{Reformulation} \citep{anthropicApril2024normalizationupdate} replaces $\mathcal{L}_P$ with the decoder-norm-weighted penalty:
\begin{equation*}
    \mathcal{L}_{P} = \sum_{i} \mathbf{f}_{i}(\mathbf{x})\, \norm{W^{dec}_{\cdot, i}}_2.
\end{equation*}
This is invariant to the homogeneous rescaling and requires no constrained optimizer. We adopt this formulation throughout. Understanding this symmetry is essential: in Section~\ref{sec: fixing} we exploit it to derive aligned training.

\subsection{SAE Architectures}
Since the original ReLU SAE \citep{huben2023sparse, bricken2023monosemanticity}, several improvements have been proposed: Gated SAEs \citep{gatedsae2024}, TopK SAEs \citep{gao2025scaling}, JumpReLU SAEs \citep{rajamanoharan2025jumping}, and BatchTopK \citep{bussmann2024batchtopk}. We primarily analyze ReLU SAEs, but demonstrate that aligned training is architecture-agnostic by showing improvements for TopK and BatchTopK variants.

\subsection{Feature Quality, Dead Features, and Stability}
A common thread runs through three persistent problems in SAE training: the 
optimization landscape is underconstrained, and gradient descent can exploit 
the resulting degeneracy in ways that hurt interpretability.

\textit{Feature quality} is most commonly measured by maximal cosine similarity 
(MCS) \citep{sharkey2023sae}, which trains two SAEs and checks which features are 
rediscovered by both; such features are considered ``universal.'' MCS scores follow 
a bimodal distribution \citep{cunningham2023replication, huben2023bimodal} and 
correlate positively with monosemanticity \citep{bills2023language, 
riggs2023latexfeature, cunningham2023saebeatsalternatives}. However, MCS is 
expensive to compute, requiring an additional larger dictionary.

\textit{Dead features} (features that never activate) are a direct symptom 
of the same degeneracy: zeroing a feature is an easy way to reduce the sparsity 
penalty at no reconstruction cost. Existing remedies (resampling 
\citep{bricken2023monosemanticity}, ghost gradients \citep{anthropic2024ghostgrads}, 
auxiliary losses \citep{gao2025scaling}) treat the symptom after the fact, each 
adding hyperparameters or non-differentiable operations. \citet{marks2024enhancingneuralnetworkinterpretability} 
take a different approach, training two SAEs in parallel with a similarity penalty 
to encourage consistent feature learning, at the cost of doubling the training budget.

\textit{Feature consistency} is a complementary concern: SAEs trained on the same 
data with different random seeds often converge to qualitatively different 
dictionaries \citep{paulo2026sparse, song2025position}, undermining reproducibility. 
This too reflects the underconstrained landscape: multiple basins of equal loss 
exist, and different seeds fall into different ones.
Archetypal SAEs \citep{fel2025archetypal} have been proposed as a stability
intervention, but in concurrent work \citep{brzozowski2026ablatingarchetypesstabilityarchetypal}
we show that their reported stability is an artifact of shared k-means
initialization rather than a genuine convergence property of the archetypal constraint.

Aligned training addresses all three problems at the source by constraining the
optimization manifold directly, rather than patching each symptom individually.

%-----------------------------------------------------------------------
\section{Proposed Methods}
%-----------------------------------------------------------------------

\subsection{Alignment Score}

According to the \textbf{linear representation hypothesis}, deep learning models represent concepts as directions in activation space. Each SAE feature $i$ corresponds to the encoder row $W^{enc}_{i,\cdot}$ and the decoder column $W^{dec}_{\cdot,i}$. It is not immediately clear which is the ``true'' feature direction; in practice the encoder is used for concept detection and the decoder for model steering \citep{wu2025axbench}. Intuitively these two directions should agree, and tied-weight SAEs \citep{huben2023sparse} enforce this as a hard constraint.

We introduce a softer proxy: the \textbf{alignment score}
\begin{equation} \label{eq:alignment_score}
    a_{i} = W^{enc}_{i,\cdot} \cdot W^{dec}_{\cdot,i},
\end{equation}
the $i$-th diagonal of $W^{enc}W^{dec}$. We use the inner product rather than cosine similarity because $a_i$ is invariant under the homogeneous rescaling of Section~\ref{background} (multiplying the encoder and dividing the decoder by any scalar leaves $a_i$ unchanged), matching the symmetry of the training objective. Cosine similarity, being separately scale-invariant in each argument, would not capture this structure.

We conjecture that features with $a_i \approx 0$ or $a_i < 0$ are uninterpretable artifacts of training. This is corroborated empirically: the alignment score follows a bimodal distribution across all modern SAE architectures, strongly correlates with MCS (Pearson $r = 0.65$), and is positively correlated with autointerpretability (Pearson $r = 0.32$). Full characterization is in Appendix~\ref{apx: bimodality}.

\paragraph{Toy model.}\label{par: toy} Consider the minimal setting: $n = 2$, $m = 1$, 
no biases, a single training point $\mathbf{x} \in \mathbb{R}^2$, and no 
sparsity penalty. Perfect reconstruction $\hat{\mathbf{x}} = \mathbf{x}$ requires
\[
    \mathbf{x} = \mathrm{ReLU}(w^{enc} \cdot \mathbf{x})\, w^{dec}.
\]
Since $\mathbf{x}$ and $w^{dec}$ must be parallel, writing $\mathbf{x} = \alpha w^{dec}$ 
and substituting yields $w^{enc} \cdot w^{dec} = 1$: perfect reconstruction forces 
the alignment score to one.

\subsection{Aligned Training} \label{sec: fixing}

The alignment score motivates a direct fix: if well-aligned features ($a_i = 1$) are the good ones, we can reparameterize the encoder to enforce this for every feature by construction. Geometrically, the constraint $W^{enc}_{i,\cdot} \cdot W^{dec}_{\cdot,i} = 1$ defines an \emph{affine hyperplane} in the space of encoder rows, with normal vector $W^{dec}_{\cdot,i}$. We parameterize the encoder by projecting a free parameter onto this hyperplane.

\paragraph{Reparameterization.} Let $A \in \mathbb{R}^{m \times n}$ be an unconstrained trainable matrix. Define each encoder row as the projection of $A_{i,\cdot}$ onto the hyperplane $\{v : v \cdot W^{dec}_{\cdot,i} = 1\}$:

\begin{equation} \label{transform}
     W^{enc}_{i,\cdot} := A_{i,\cdot} + \alpha_i\, W^{dec}_{\cdot,i},
     \qquad
     \alpha_i = \frac{1 - A_{i,\cdot} \cdot W^{dec}_{\cdot,i}}{\norm{W^{dec}_{\cdot,i}}^{2}}.
\end{equation}

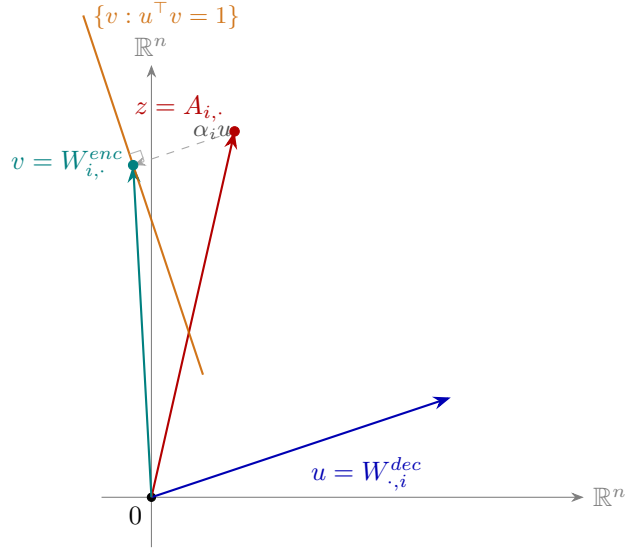
\begin{figure}[t]
\centering
\begin{tikzpicture}[scale=2.2, >=Stealth]
    % Axes
    \draw[->, thin, gray] (-0.3,0) -- (2.6,0) node[right, gray] {$\mathbb{R}^n$};
    \draw[->, thin, gray] (0,-0.3) -- (0,2.6) node[above, gray] {$\mathbb{R}^n$};
    \filldraw[black] (0,0) circle (0.7pt) node[below left] {$0$};

    % Decoder direction u (normal to hyperplane)
    \coordinate (U) at (1.8, 0.6);
    \draw[->, thick, blue!70!black] (0,0) -- (U)
        node[midway, below right, blue!70!black] {$u = W^{dec}_{\cdot,i}$};

    % Free parameter z = A_{i,.}
    \coordinate (Z) at (0.5, 2.2);
    \draw[->, thick, red!70!black] (0,0) -- (Z)
        node[above left, red!70!black] {$z = A_{i,\cdot}$};

    % Projected point v = W^enc_{i,.}  (lies on hyperplane)
    % Hyperplane {v : u^T v = 1}. With u=(1.8,0.6), ||u||^2=3.6.
    % u^T z = 1.8*0.5+0.6*2.2 = 0.9+1.32=2.22
    % alpha = (1-2.22)/3.6 = -1.22/3.6 = -0.3389
    % v = z + alpha*u = (0.5 - 0.3389*1.8, 2.2 - 0.3389*0.6)
    %                 = (0.5-0.610, 2.2-0.2033) = (-0.11, 1.997)
    \coordinate (V) at (-0.11, 1.997);
    \draw[->, thick, teal] (0,0) -- (V)
        node[left, teal] {$v = W^{enc}_{i,\cdot}$};

    % Shift arrow alpha*u from z to v
    \draw[->, dashed, gray!70] (Z) -- (V)
        node[midway, above right, gray!60!black, font=\small] {$\alpha_i u$};

    % Hyperplane line {v : u^T v = 1}
    % Normal u=(1.8,0.6). A point on the plane: p = u/||u||^2 = (0.5, 0.1667)
    % Tangent direction t = (-0.6, 1.8) (perp to u)
    % let's just use V as anchor: V + s*T
    \coordinate (HL_a) at ({-0.11 + (-0.7)*(-0.6)}, {1.997 + (-0.7)*(1.8)});
    \coordinate (HL_b) at ({-0.11 + (0.5)*(-0.6)},  {1.997 + (0.5)*(1.8)});
    \draw[thick, orange!80!black, opacity=0.9] (HL_a) -- (HL_b)
        node[right, orange!80!black, font=\small] {$\{v : u^\top v = 1\}$};

    % Dot on hyperplane at V
    \filldraw[teal] (V) circle (0.8pt);
    % Dot on z
    \filldraw[red!70!black] (Z) circle (0.8pt);

    % Right-angle marker at V showing alpha*u perp to hyperplane
    % tangent direction (normalised): t_norm = (-0.6,1.8)/|(−0.6,1.8)| = (-0.6,1.8)/1.897
    \def\eps{0.07}
    \coordinate (sq1) at ({-0.11 + \eps*1.8/1.897}, {1.997 + \eps*0.6/1.897});
    \coordinate (sq2) at ({-0.11 + \eps*1.8/1.897 + \eps*(-0.6)/1.897},
                           {1.997 + \eps*0.6/1.897  + \eps*1.8/1.897});
    \coordinate (sq3) at ({-0.11 + \eps*(-0.6)/1.897}, {1.997 + \eps*1.8/1.897});
    \draw[gray!60, thin] (sq1) -- (sq2) -- (sq3);
\end{tikzpicture}
\caption{Geometric interpretation of aligned training for a single feature $i$. The free parameter $z = A_{i,\cdot}$ is projected onto the affine hyperplane $\{v : u^\top v = 1\}$ (orange line) by adding a scalar multiple $\alpha_i u$ of the decoder direction $u = W^{dec}_{\cdot,i}$. The result $v = W^{enc}_{i,\cdot}$ satisfies the alignment constraint by construction. The small square marks the right angle between $\alpha_i u$ and the hyperplane.}
\label{fig:projection}
\end{figure}

One verifies immediately that $W^{enc}_{i,\cdot} \cdot W^{dec}_{\cdot,i} = 1$ for every $i$. The biases $\mathbf{b}^{enc}$, $\mathbf{b}^{dec}$ and the decoder $W^{dec}$ remain unrestricted trainable parameters; standard gradient descent is applied to $(A, \mathbf{b}^{enc}, \mathbf{b}^{dec}, W^{dec})$. Figure~\ref{fig:projection} illustrates the geometry.

\paragraph{Derivation.} The constraint $u^\top v = 1$ (with $u = W^{dec}_{\cdot,i}$, $v = W^{enc}_{i,\cdot}$) defines an affine hyperplane with normal $u$. Any point on this hyperplane can be written as $v = z + \alpha u$ for a free parameter $z \in \mathbb{R}^n$. Substituting into the constraint:
\[
    u^\top(z + \alpha u) = 1
    \quad\Longrightarrow\quad
    \alpha = \frac{1 - u^\top z}{\norm{u}^2}.
\]
Equivalently, every solution decomposes as
\begin{equation} \label{eq:hyperplane}
    v = \frac{u}{\norm{u}^2} + \underbrace{\left(I - \frac{uu^\top}{\norm{u}^2}\right)}_{\text{proj onto } u^\perp} z,
\end{equation}
a fixed particular solution $u/\norm{u}^2$ plus a component lying in the null space of $u^\top$ (the tangent space of the hyperplane). Equation~\eqref{transform} is exactly this formula. An alternative derivation via the Moore-Penrose pseudoinverse is given in Appendix~\ref{apx: projection}.

\paragraph{Compression for free.} Since equation~\eqref{transform} constrains each encoder row to lie on an affine hyperplane of dimension $n-1$, the last element of each row of $A$ can be fixed to zero and left untrained. The aligned SAE therefore has slightly \emph{fewer} parameters than the standard SAE while expressing the same space of reachable weights.

\paragraph{Connection to inductive biases.} This approach is analogous to convolutional networks: a convolutional layer is a linear layer constrained by local receptive fields and weight sharing \citep{FUKUSHIMA1988119, LeNet}. In principle a linear layer could learn convolution, but the inductive bias makes training more effective. Here, aligned training constrains the encoder to a manifold that concentrates probability on the good basin of the optimization landscape.

%-----------------------------------------------------------------------
\section{Experiments}
%-----------------------------------------------------------------------

\subsection{Setup} \label{sec:setup}

We train ReLU, TopK \citep{gao2025scaling}, and BatchTopK \citep{bussmann2024batchtopk} sparse autoencoders on 50 million tokens from The Pile \citep{pile}, evaluated on OpenWebText. The primary evaluation models are layer 8 of Pythia 160M and layer 12 of Gemma 2 2B \citep{gemmateam2024gemma2improvingopen}, following the SAEBench protocol \citep{karvonen2025saebenchcomprehensivebenchmarksparse}. We train at dictionary sizes 4K, 16K, and 65K; for ReLU SAEs we sweep four sparsity penalties $\lambda$. All comparisons are across matched sparsity levels measured by $L_0$. Full model and hyperparameter details are in Appendix~\ref{apx: model}.

\subsection{Reconstruction Metrics}

\paragraph{Metrics.} Reconstruction is evaluated by explained variance and recovered cross-entropy loss at varying sparsity levels. Explained variance is
\begin{equation*}
    1 - \frac{\frac{1}{K}\sum_{k}||\mathbf{x}_k - \hat{\mathbf{x}}_k||^{2}}{\frac{1}{K}\sum_{k}||\mathbf{x}_k - \bm{\mu}||^{2}},
\end{equation*}
where $\bm{\mu}$ is the mean activation. Recovered cross-entropy is
\begin{equation*}
\frac{H^* - H_0}{H_{orig} - H_0},
\end{equation*}
with $H_{orig}$ the model's next-token cross-entropy, $H^*$ the cross-entropy when activations are replaced by SAE reconstructions, and $H_0$ the cross-entropy when activations are zeroed.

\paragraph{Results.} Aligned training achieves Pareto improvements over standard training across both models, all three dictionary sizes, and all four sparsity levels (Figure~\ref{fig:reconstruction}; see also Figures~\ref{fig:reconstruction 16K}--\ref{fig:reconstruction 65 K} in Appendix~\ref{apx: additional}).

\begin{figure}[htb!]
    \centering
    \includegraphics[width=1.0\linewidth]{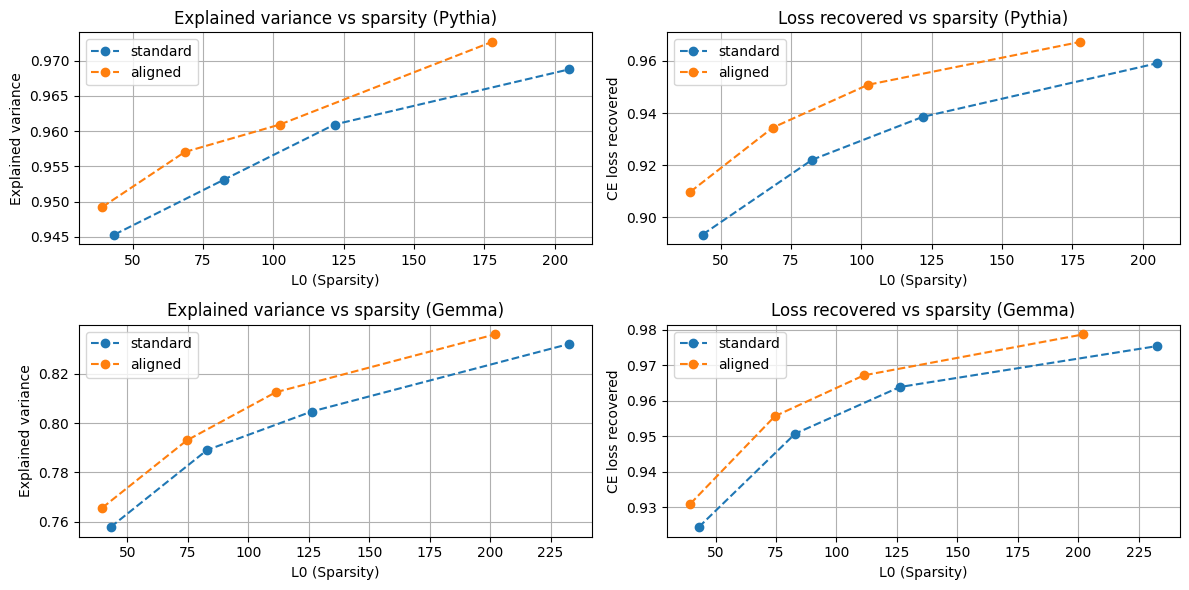}
    \caption{Aligned training improves recovered cross-entropy across different sparsity levels. Dictionary size 4096, layer 8 of Pythia 160M and layer 12 of Gemma 2 2B.}
    \label{fig:reconstruction}
\end{figure}

We further validate this at scale: Appendix~\ref{apx: 500M} scales training to 500M tokens from The Pile and compares against the SAEBench state-of-the-art checkpoints, where aligned training continues to outperform on reconstruction metrics.

\begin{figure}[htb!]
    \centering
    \includegraphics[width=0.8\linewidth]{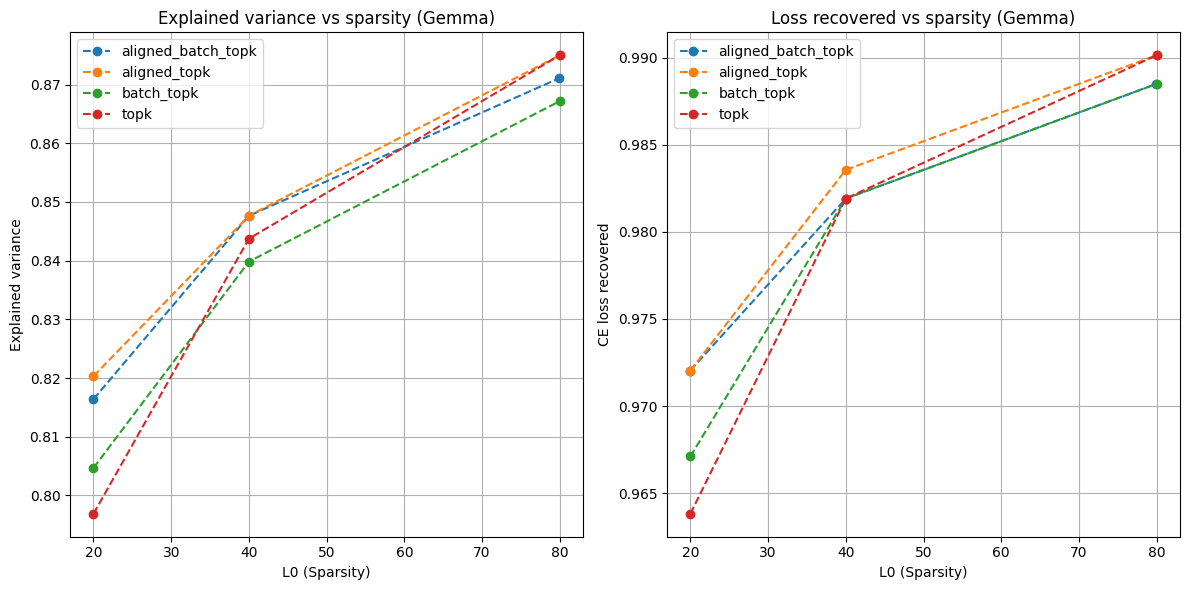}
    \caption{Aligned training improves TopK and BatchTopK autoencoders in the low-sparsity regime. Dictionary size 65K, layer 12 of Gemma 2 2B.}
    \label{fig:reconstruction top k}
\end{figure}

\paragraph{TopK and BatchTopK.} Aligned training extends directly to TopK and BatchTopK architectures, where sparsity is enforced by the top-$k$ activation rather than an $L^1$ penalty. Figure~\ref{fig:reconstruction top k} shows improvements in the low-sparsity regime for the 65K Gemma dictionary. Results are consistent across dictionary sizes.

\subsection{Dead Features}

Dead features --- features that never activate --- are a persistent training pathology. Standard approaches address them after the fact: resampling \citep{bricken2023monosemanticity}, ghost gradients \citep{anthropic2024ghostgrads}, and auxiliary losses \citep{gao2025scaling} all add hyperparameters or non-differentiable operations. Standard ReLU SAEs exhibit approximately 20\% dead features at dictionary size 4K.

\begin{figure}[htb!]
    \centering
    \includegraphics[width=1.0\linewidth]{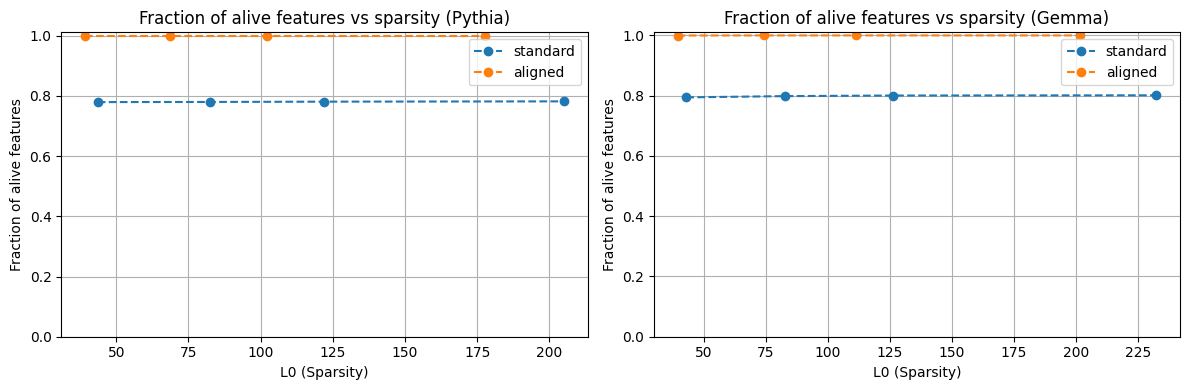}
    \caption{Aligned training reduces dead features to near zero without resampling or auxiliary losses. Dictionary size 4096, layer 8 of Pythia 160M and layer 12 of Gemma 2 2B.}
    \label{fig:dead neurons}
\end{figure}

\begin{figure}[htb!]
    \centering
    \includegraphics[width=0.8\linewidth]{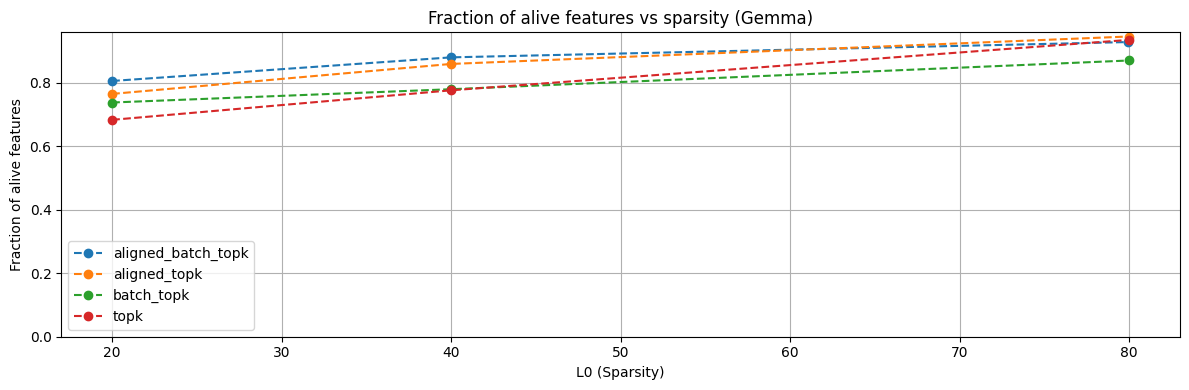}
    \caption{The dead-feature reduction extends to TopK and BatchTopK. Dictionary size 65K, layer 12 of Gemma 2 2B.}
    \label{fig:dead neurons top k}
\end{figure}

Aligned training eliminates dead features structurally rather than resurrect them after they die. By constraining the optimization manifold, it prevents features from entering the zero-activation regime in the first place. Figure~\ref{fig:dead neurons} shows that aligned training reduces dead features to near zero across both Pythia and Gemma, at no additional cost. The same effect holds for TopK and BatchTopK (Figure~\ref{fig:dead neurons top k}) and at 500M-token scale (Appendix~\ref{apx: 500M}).

\subsection{Stability Across Seeds}

\paragraph{Motivation.} A natural question is whether two SAEs trained independently on the same data, same architecture, and same hyperparameters but different random seeds learn similar dictionaries. Recent work \citep{paulo2026sparse, song2025position} shows the answer is often no: standard SAEs exhibit substantial instability, with independently trained models converging to qualitatively different dictionaries. This undermines reproducibility in mechanistic interpretability.

\paragraph{Metric.} We measure stability as the mean min cosine similarity (MMCS) between the decoder dictionaries of two independently trained SAEs: for each feature in the first SAE we find its nearest neighbor in the second and average the cosine similarities. A score of 1 indicates perfect agreement; lower scores indicate divergent dictionaries.

\paragraph{Why alignment helps.} Standard SAEs have two sources of instability beyond the inherent permutation symmetry: (i) the homogeneous rescaling degeneracy of Section~\ref{background}, and (ii) the independent encoder and decoder, which introduce rotational degrees of freedom in the optimization landscape. Aligned training removes (ii) directly: the constraint $W^{enc}_{i,\cdot} \cdot W^{dec}_{\cdot,i} = 1$ couples each encoder row to its decoder column, shrinking the manifold of reachable weights and biasing optimization toward a more consistent basin.

\begin{figure}[htb!]
    \centering
    \includegraphics[width=1.0\linewidth]{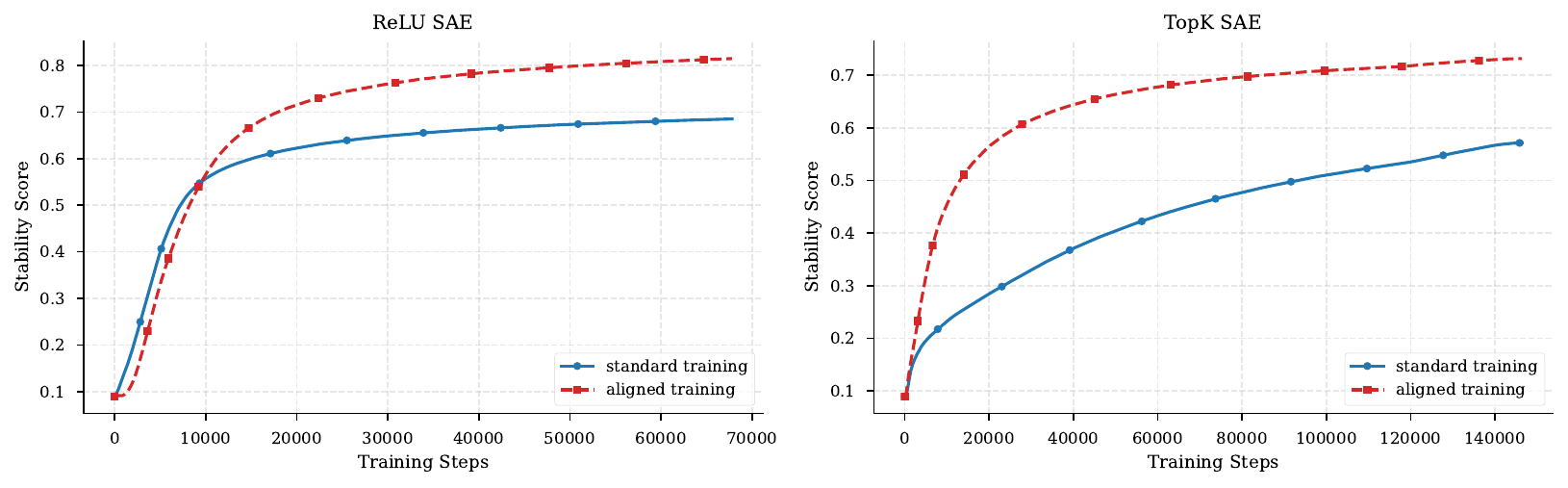}
    \caption{Aligned training significantly improves cross-seed stability for both ReLU and TopK autoencoders. Dictionary size 65K, layer 12 of Gemma 2 2B.}
    \label{fig:stability combined}
\end{figure}

\paragraph{Results.} Figure~\ref{fig:stability combined} shows MMCS as a function of training progress for standard and aligned TopK SAEs (Gemma 2 2B, layer 12, dictionary size 65K, $k=20$). Aligned TopK achieves consistently lower cosine distance throughout training, and the gap opens early, demonstrating the effect is not merely an initialization artifact. Archetypal SAEs \citep{fel2025archetypal} have also been proposed to address instability, though in concurrent work \citep{brzozowski2026ablatingarchetypesstabilityarchetypal} we demonstrate that their stability gains are an artifact of shared initialization rather than a property of the archetypal constraint itself.

% brought from app.
\subsection{Combined with p-Annealing} \label{app:p-annealing}

The $L^1$ sparsity penalty in the standard SAE is a convex relaxation of the 
intractable $L^0$ norm, and is known to induce \textit{feature shrinkage}: 
active features are systematically underestimated in magnitude because the 
penalty penalizes their activation values directly \citep{wright2024shrinkage}. 
p-Annealing \citep{karvonen2024measuring} addresses this by replacing the $L^1$ 
penalty with a smooth $L^p$ quasinorm ($0 < p \leq 1$) and annealing $p$ toward 
zero during training. Since $L^p$ for $p < 1$ is nonconvex but smooth, it 
provides a closer approximation to $L^0$ while remaining differentiable, and at 
inference the resulting model is architecturally identical to a standard ReLU SAE. 
Feature shrinkage and the encoder-decoder misalignment we identify are 
orthogonal failure modes, so the two methods compose without interference.

\begin{figure}[htbp!]
    \centering
    \includegraphics[width=0.7\linewidth]{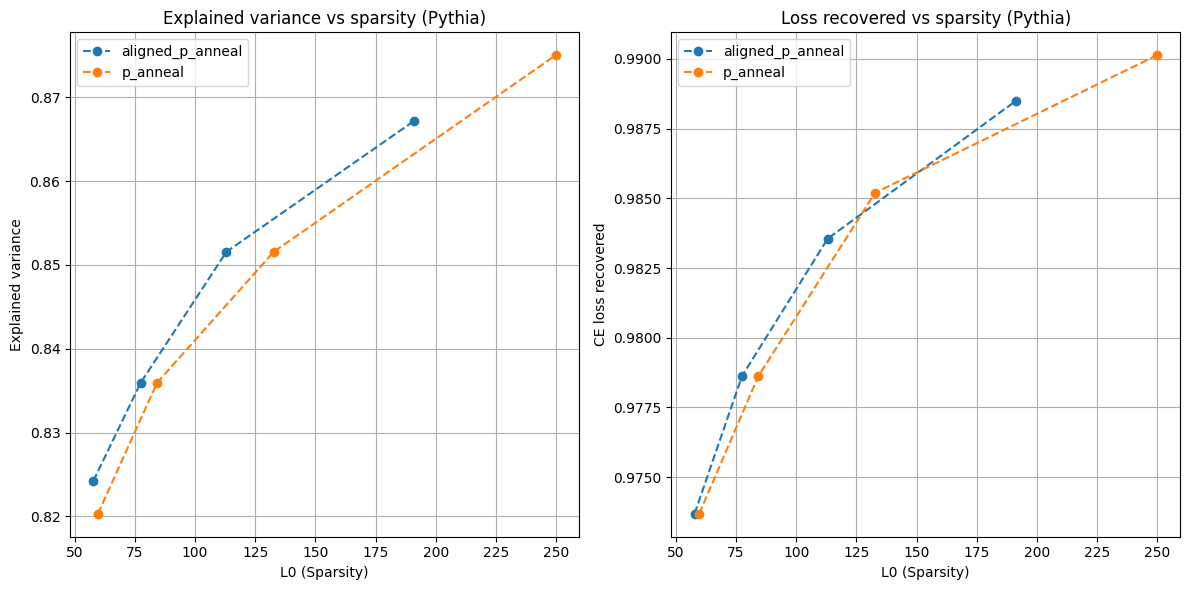}
    \caption{Reconstruction metrics for Pythia 160M (layer 8), dictionary size 4096, 3 random seeds.}
    \label{fig: reconstruction p anneal}
\end{figure}

\begin{figure}[htbp!]
    \centering
    \includegraphics[width=0.7\linewidth]{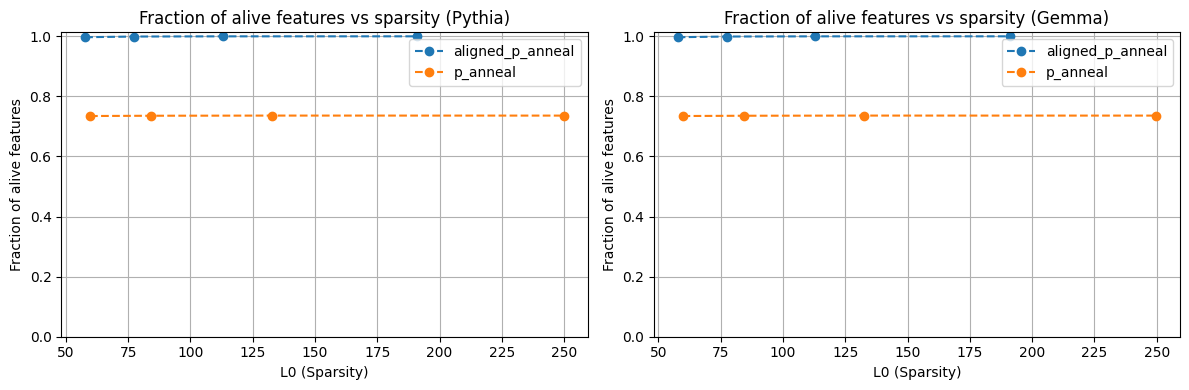}
    \caption{Alive-feature fraction for Pythia 160M (layer 8), dictionary size 4096.}
    \label{fig: dead neurons p anneal}
\end{figure}

Figure~\ref{fig: reconstruction p anneal} and Figure~\ref{fig: dead neurons p anneal} 
compare p-annealing alone against aligned training combined with p-annealing on 
Pythia 160M (layer 8), dictionary size 4096. On dead features, the combination 
reduces the dead fraction to near zero across all sparsity levels, whereas 
p-annealing alone plateaus at roughly 73\% alive features (Figure~\ref{fig: dead 
neurons p anneal}). On reconstruction, the combination achieves higher explained 
variance across most sparsity levels, with p-annealing alone showing a marginal 
advantage on CE loss recovered only at the highest sparsity levels 
(Figure~\ref{fig: reconstruction p anneal}). Overall, aligned training provides a 
consistent improvement over p-annealing alone, particularly on feature 
utilization.

% \subsection{Stability Analysis Across Seeds} \label{app:seeds}

% Figure~\ref{seeds} shows reconstruction metrics across 3 random seeds for aligned training, confirming that the reported improvements are stable and not driven by seed-specific luck.

% \begin{figure}[htbp!]
%     \centering
%     \includegraphics[width=0.8\linewidth]{figures/rebuttal/reconstruction_seed_4K.png}
%     \caption{Reconstruction metrics for Pythia 160M (layer 8), dictionary size 4096, 3 random seeds.}
%     \label{seeds}
% \end{figure}

\subsection{Spurious Correlation Removal} \label{app:spurious}

Reconstruction quality is a proxy metric: it measures how faithfully the SAE 
reproduces activations, but not whether the learned features are useful for 
downstream interpretability tasks. To assess the latter, we evaluate on the 
Spurious Correlation Removal (SCR) benchmark from SAEBench 
\citep{karvonen2025saebenchcomprehensivebenchmarksparse}, a feature 
disentanglement task adapted from the SHIFT method 
\citep{marks2025sparse, karvonen2024evaluatingsparseautoencoderstargeted}.

In SCR, a classifier is trained to predict a target concept (e.g., profession) 
from SAE latents, where the training data contains a spurious correlation with 
a confounding attribute (e.g., gender). SAE latents identified as encoding 
the spurious signal are zero-ablated, and the resulting classifier accuracy on 
a balanced held-out set measures how cleanly the SAE has disentangled the two 
concepts. A higher SCR score indicates that the SAE has isolated the spurious 
attribute into dedicated latents, making it surgically removable without 
collateral damage to the target concept.

\begin{figure}[htb!]
    \centering
    \includegraphics[width=1.0\linewidth]{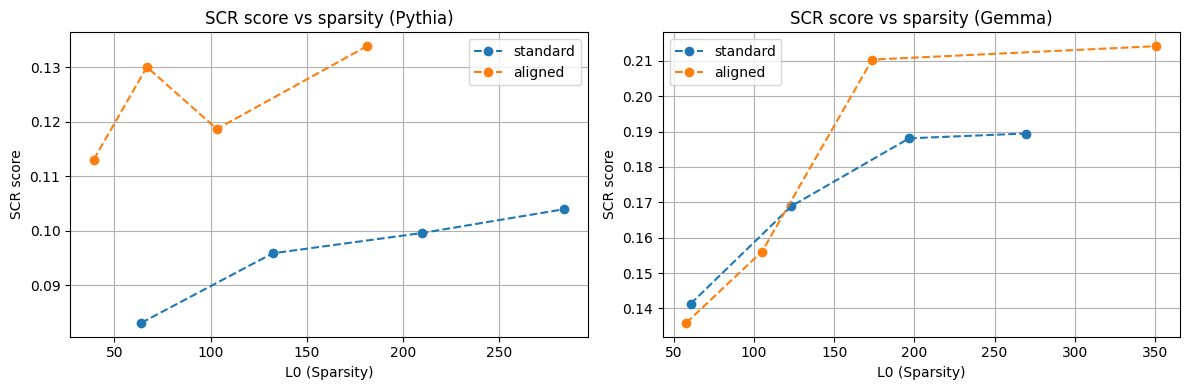}
    \caption{SCR metric from SAEBench, dictionary size 65K, Pythia 160M and Gemma 2 2B.}
    \label{fig: scr}
\end{figure}

Figure~\ref{fig: scr} shows SCR scores at dictionary size 65K for Pythia 160M 
and Gemma 2 2B. On Pythia, aligned training outperforms the standard baseline 
consistently across all sparsity levels. On Gemma the picture is more nuanced: 
standard training has a marginal advantage at very low sparsity ($L_0 \leq 110$), after which aligned training overtakes it and the gap widens steadily 
with increasing $L_0$. This pattern suggests that the benefits of alignment 
for feature disentanglement become more pronounced as the SAE is forced to be 
more selective about which features it activates.

\section{Limitations}
\label{sec:limitations}
Limitations. Our empirical evaluation focuses on two model families (Pythia 160M and Gemma 2 2B) and a fixed set of SAE architectures. Whether the benefits of aligned training persist at larger model scales or under substantially different training regimes remains an open question for future work.

%-----------------------------------------------------------------------
\section{Conclusion}
%-----------------------------------------------------------------------

Despite the popularity and success of SAE in understanding the activations in DNNs, high dimensional features extracted from SAE exhibit critical challenges such as instability and dead features. To improve feature quality in a wide range of models without incurring additional training or data, we introduce the aligned training.

The alignment score is the inner product between each SAE feature's encoder row and decoder column, which are shown to follow a bimodal distribution in standard SAEs and strongly correlates with existing quality proxies. The proposed aligned training enforces the alignment score to be 1 for every feature via a parameter-free reparameterization: the encoder is defined as the projection of a free parameter matrix $A$ onto the affine hyperplane $\{v : v \cdot W^{dec}_{\cdot,i} = 1\}$.

This single geometric constraint, with no additional hyperparameters, simultaneously produces three improvements: Pareto improvements in reconstruction quality, elimination of dead features, and significantly more stable dictionaries across seeds. These improvements are observed across different models, dictionary sizes, sparsity levels, and SAE architectures. Training budgets of 50M and 500M tokens resulted in similar improvements. Furthermore, we have shown that other training techniques for SAE such as p-Annealing can be used together with the aligned training.

The stability result directly addresses the feature consistency concerns raised 
by \citet{paulo2026sparse} and \citet{song2025position} as a barrier to reliable 
mechanistic interpretability. \citet{paulo2026sparse} show that in large SAEs, 
as few as 30\% of features are shared across independent training runs, and that 
this instability is particularly pronounced for TopK architectures. 
\citet{song2025position} argue that consistency should be treated as a primary 
design objective rather than an afterthought. Aligned training addresses this 
directly: by coupling each encoder row to its decoder column through the 
constraint $W^{enc}_{i,\cdot} \cdot W^{dec}_{\cdot,i} = 1$, it removes a 
rotational degree of freedom that allows different seeds to converge to 
geometrically distinct but equally valid dictionaries. As shown in 
Figure~\ref{fig:stability combined}, the MMCS gap between aligned and standard 
TopK SAEs opens early in training and widens throughout, indicating the 
improvement is structural rather than an artifact of initialization.

\bibliography{bibliography}
\bibliographystyle{plainnat}

%%%%%%%%%%%%%%%%%%%%%%%%%%%%%%%%%%%%%%%%%%%%%%%%%%%%%%%%%%%%
\appendix

\section{Implementation Details}
\label{apx: details}

All SAEs are trained on 50 million tokens from The Pile, 
evaluated on OpenWebText, following the SAEBench protocol. 
The primary evaluation models are layer 8 of Pythia 160M 
(residual stream dimension 768, \texttt{float32}) and layer 
12 of Gemma 2 2B (residual stream dimension 2304, 
\texttt{bfloat16}). All experiments are conducted on a 
single NVIDIA H100 80GB GPU. A single SAE training run completes in approximately 13 minutes on this hardware.

Training uses the Adam optimizer with learning rate 
$3 \times 10^{-4}$, a linear warmup over 1000 steps, 
sparsity penalty warmup over 5000 steps, and learning rate 
decay beginning at 80\% of total training steps. The SAE 
batch size is 2048 for all models. Dictionary sizes are 4K, 
16K, and 65K. For ReLU SAEs, we sweep four sparsity 
penalties $\lambda \in \{0.025, 0.035, 0.045, 0.07\}$.

\section{Code Implementation} \label{apx: code}

For training our autoencoders, we utilized the code \citep{marks2024dictionary_learning} and implemented the key transform~\eqref{transform} using the Python function below. For computing per-feature inner products we used the einops \citep{rogozhnikov2022einops} library.

\begin{lstlisting}
def get_the_encoder_matrix(
    dict_size: int,
    encoder_weights_orthogonal_part: Float[Tensor, "activation_dim-1 dict_size"],
    decoder_weights: Float[Tensor, "dict_size activation_dim"]
) -> Float[Tensor, "activation_dim dict_size"]:
    zeros = torch.zeros(1, dict_size).to(encoder_weights_orthogonal_part)
    appended = torch.concat([encoder_weights_orthogonal_part, zeros])
    inner_products = einops.einsum(
        decoder_weights, appended,
        "dict_size activation_dim, activation_dim dict_size -> dict_size"
    )
    decoder_norms_squared = decoder_weights.pow(2).sum(dim=1)
    reparametrized = appended + decoder_weights.T * (1 - inner_products) / decoder_norms_squared
    return reparametrized
\end{lstlisting}

\section{Alignment Score: Bimodality and Correlations} \label{apx: bimodality}

\subsection{Alignment Scores Are Bimodal}

\begin{figure}[htb!]
    \centering
    \includegraphics[width=1.0\linewidth]{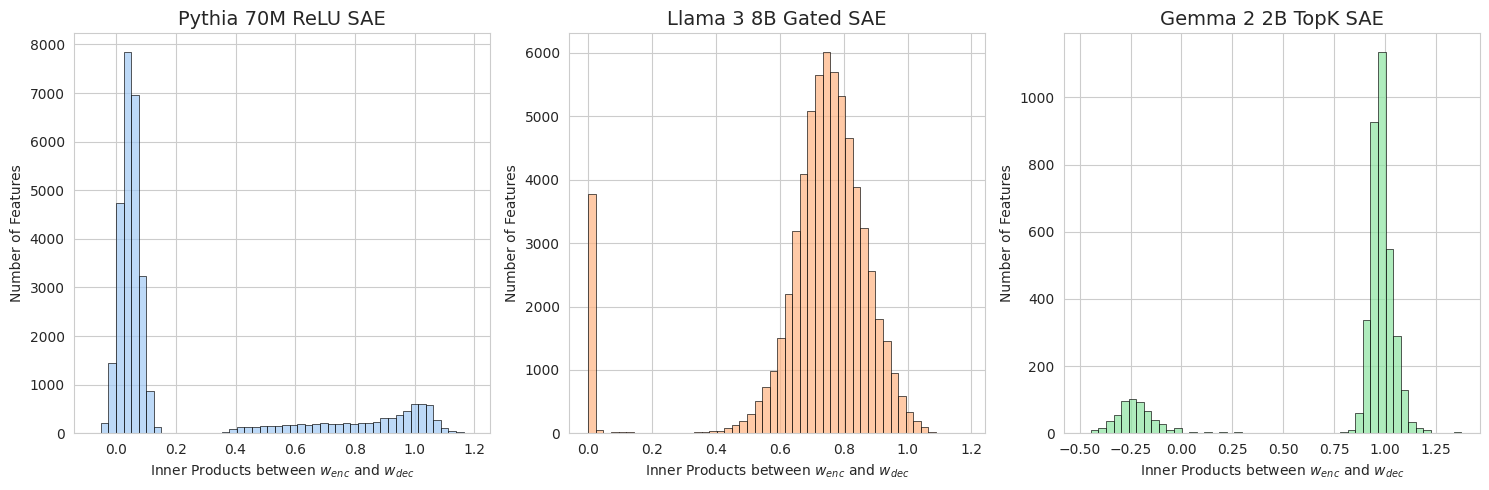}
    \caption{Bimodality of SAE alignment scores across different models and architectures.}
    \label{fig:histograms}
\end{figure}

We computed histograms of the alignment score across Pythia 70M, LLaMA 3 8B IT, and Gemma 2 2B, using standard pretrained SAEs loaded via the sae-lens framework \citep{bloom2024saetrainingcodebase}. The score distribution is consistently bimodal across all models and architectures (Figure~\ref{fig:histograms}). This phenomenon mirrors the bimodality of MCS reported by \citet{huben2023bimodal}, but the alignment score requires no additional training run. Specific SAE details are in Appendix~\ref{apx: model}.

\subsection{Alignment Scores Are Highly Correlated with MCS}

\begin{figure}[htb!]
    \centering
    \includegraphics[width=0.7\linewidth]{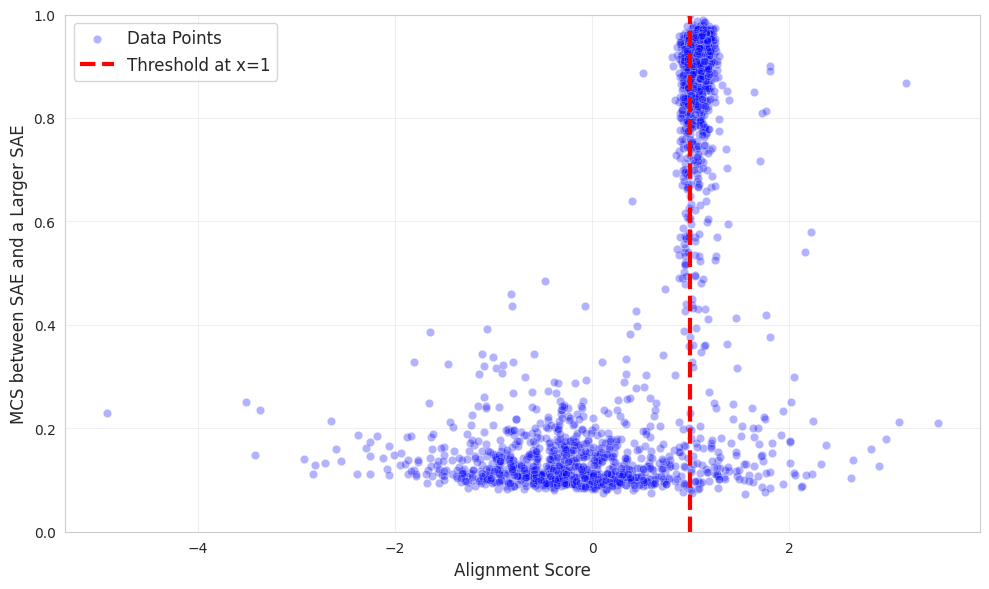}
    \caption{MCS vs.\ alignment score (Pearson $r = 0.65$). The red vertical line marks $a_i = 1$.}
    \label{fig:mcs_correlation}
\end{figure}

We compared alignment scores with MCS for an SAE trained on MLP activations from layer 2 of Pythia \citep{biderman2023pythia}, using the SAEs from \citep{riggs2023latexfeature}. Figure~\ref{fig:mcs_correlation} shows a Pearson correlation of $0.65$; low-alignment features cluster away from the diagonal and are consistently missed by the larger interpreter model. The best features concentrate near $a_i = 1$, consistent with the toy model prediction (Section~\ref{par: toy}).

\subsection{Alignment Scores Are Correlated with Autointerpretability}

\begin{figure}[htb!]
    \centering
    \includegraphics[width=0.7\linewidth]{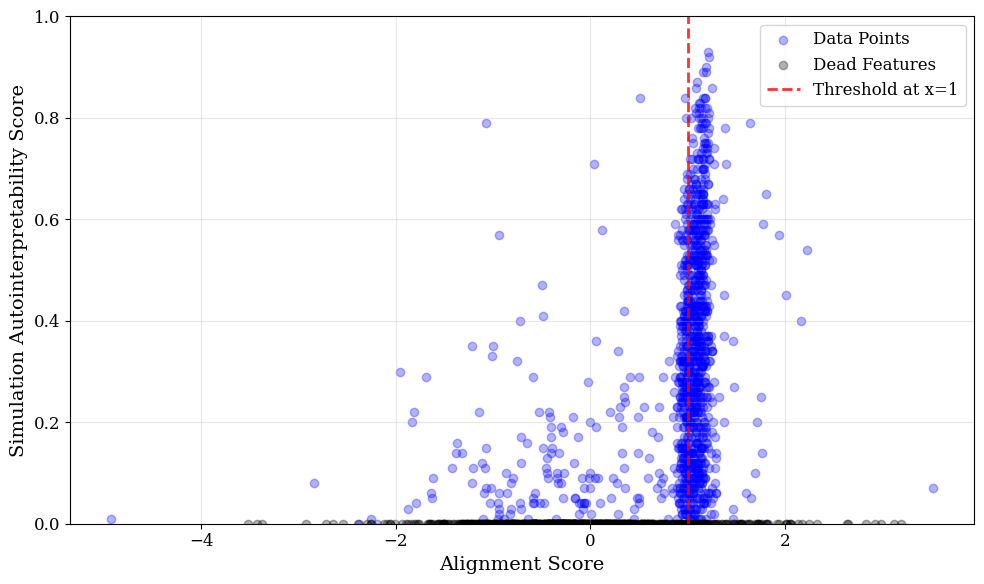}
    \caption{Autointerpretability vs.\ alignment score (Pearson $r = 0.32$). Red line marks $a_i = 1$.}
    \label{fig:autointerpretability}
\end{figure}

The alignment score is positively correlated with autointerpretability (Pearson $r = 0.32$; Figure~\ref{fig:autointerpretability}), using the protocol from \citep{bills2023language} with Gemma 3 27B IT as judge. Dead features (insufficient non-zero activations) are shown as black dots.

\section{Alternative Derivation via Pseudoinverse} \label{apx: projection}

The inline derivation in Section~\ref{sec: fixing} parameterizes the affine hyperplane $\{v : u^\top v = 1\}$ by shifting a free vector $z$ along the normal $u$. The same result follows from the Moore-Penrose pseudoinverse. The general solution of $u^\top v = 1$ is
\[
    v = (u^\top)^+ + \left(I - (u^\top)^+ u^\top\right)z, \qquad z \in \mathbb{R}^n,
\]
where for a row vector $u^\top$ the pseudoinverse is $(u^\top)^+ = u/\norm{u}^2$. Substituting yields equation~\eqref{eq:hyperplane} immediately, confirming that the two formulations are identical.

\section{Models and SAE Details} \label{apx: model}

For loading pretrained SAEs we used the sae-lens framework \citep{bloom2024saetrainingcodebase}. SAEs used in Appendix~\ref{apx: bimodality}:
\begin{itemize}
    \item Pythia: 70M, residual stream layer 3. Release: \textit{pythia-70m-deduped-res-sm}, id: \textit{blocks.3.hook\_resid\_post}.
    \item LLaMA: 3 8B IT, residual stream layer 25. Release: \textit{llama-3-8b-it-res-jh}, id: \textit{blocks.25.hook\_resid\_post}.
    \item Gemma: 2 2B, residual stream layer 12. Release: \textit{sae\_bench\_gemma-2-2b\_topk\_width-2pow12\_date-1109}, id: \textit{blocks.12.hook\_resid\_post\_\_trainer\_0}.
\end{itemize}
For the MCS and autointerpretability plots we used the ReLU autoencoders from \citep{riggs2023latexfeature}. For the remaining experiments we used layer 8 of Pythia 160M and layer 12 of Gemma 2 2B, following SAEBench.

\section{Comparison to Weight-Tying} \label{app:weight-tying}

\begin{figure}[ht!]
    \centering
    \includegraphics[width=0.7\linewidth]{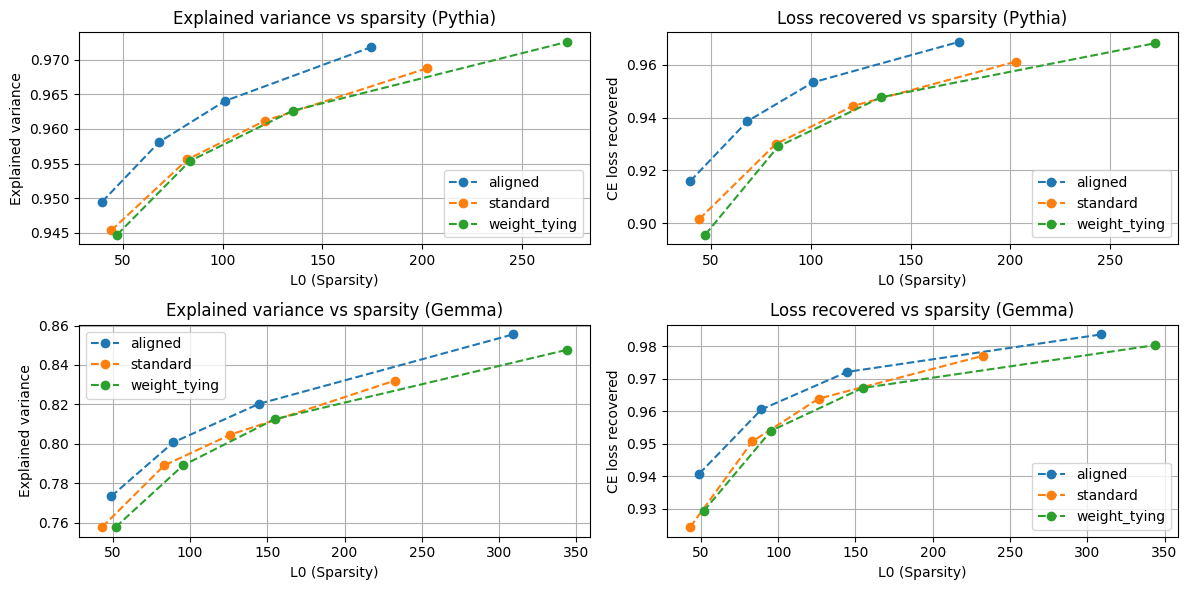}
    \caption{Reconstruction metrics for Pythia 160M (layer 8) and Gemma 2 2B (layer 12), dictionary size 16384. Weight tying underperforms on reconstruction.}
    \label{fig: reconstruction weight tying}
\end{figure}

\begin{figure}[htbp!]
    \centering
    \includegraphics[width=1.0\linewidth]{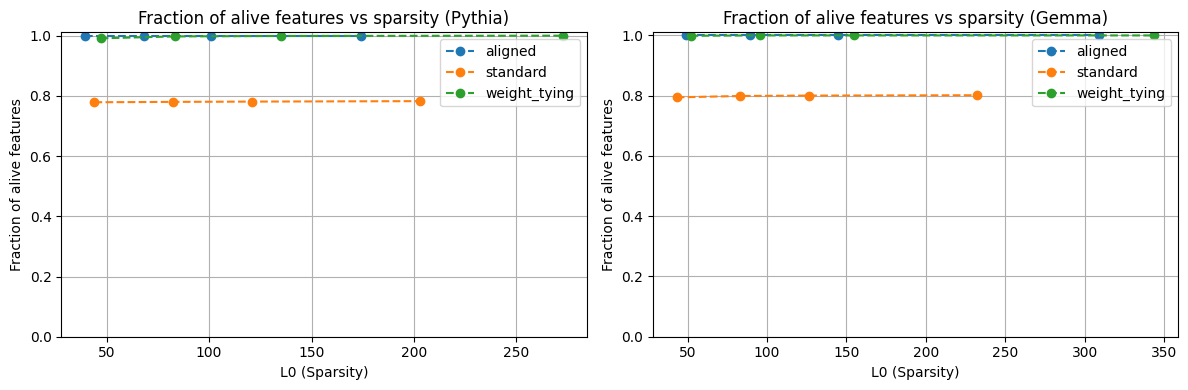}
    \caption{Weight tying reduces dead features but at the cost of reconstruction quality. Pythia 160M (layer 8) and Gemma 2 2B (layer 12), dictionary size 16384.}
    \label{fig: dead neurons weight tying}
\end{figure}

Weight tying \citep{huben2023sparse} enforces $W^{enc} = (W^{dec})^T$, a strictly stronger constraint than aligned training. Experiments show that weight tying reduces dead features but compromises reconstruction quality (Figures~\ref{fig: reconstruction weight tying}--\ref{fig: dead neurons weight tying}). Aligned training achieves improvements on both metrics simultaneously.

\newpage

\section{Results for Different Dictionary Sizes} \label{apx: additional}

\begin{figure}[htbp!]
    \centering
    \includegraphics[width=1.0\linewidth]{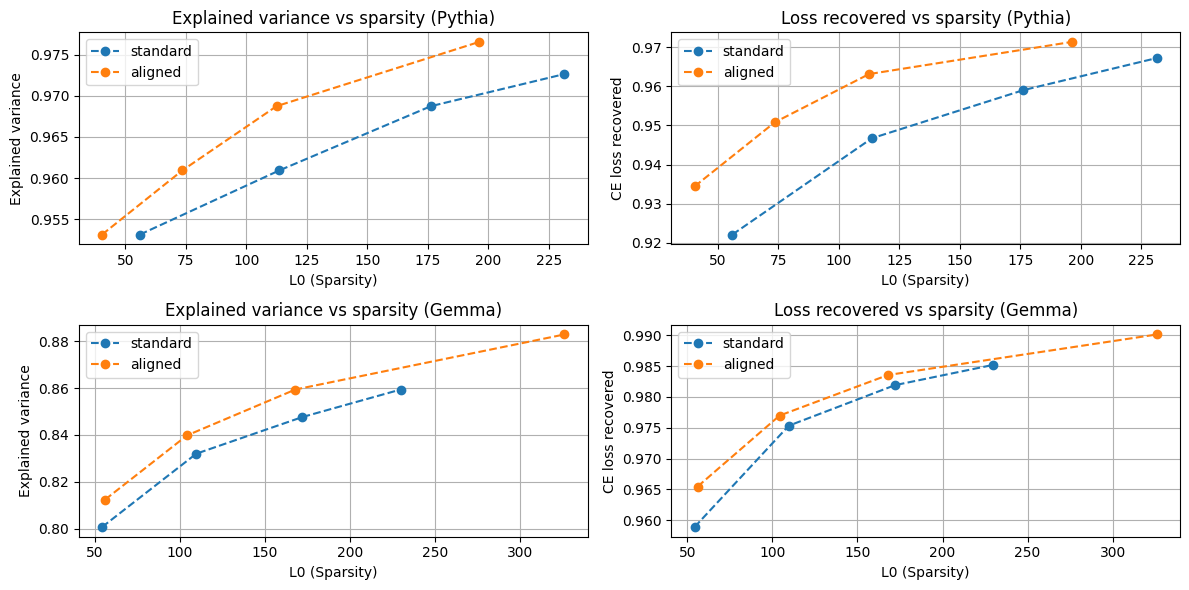}
    \caption{Reconstruction metrics, dictionary size 16384, Pythia 160M (layer 8) and Gemma 2 2B (layer 12).}
    \label{fig:reconstruction 16K}
\end{figure}

\begin{figure}[htbp!]
    \centering
    \includegraphics[width=1.0\linewidth]{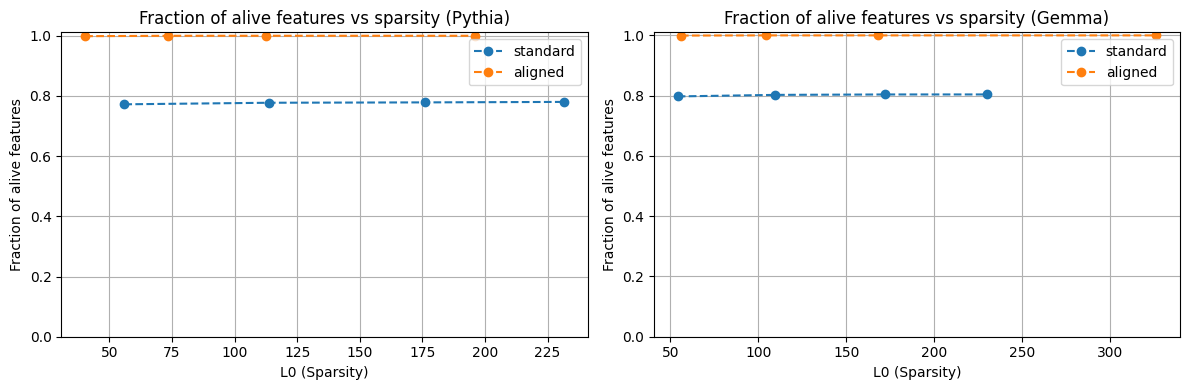}
    \caption{Alive-feature fraction, dictionary size 16384.}
    \label{fig:dead neurons 16 K}
\end{figure}

\begin{figure}[htbp!]
    \centering
    \includegraphics[width=1.0\linewidth]{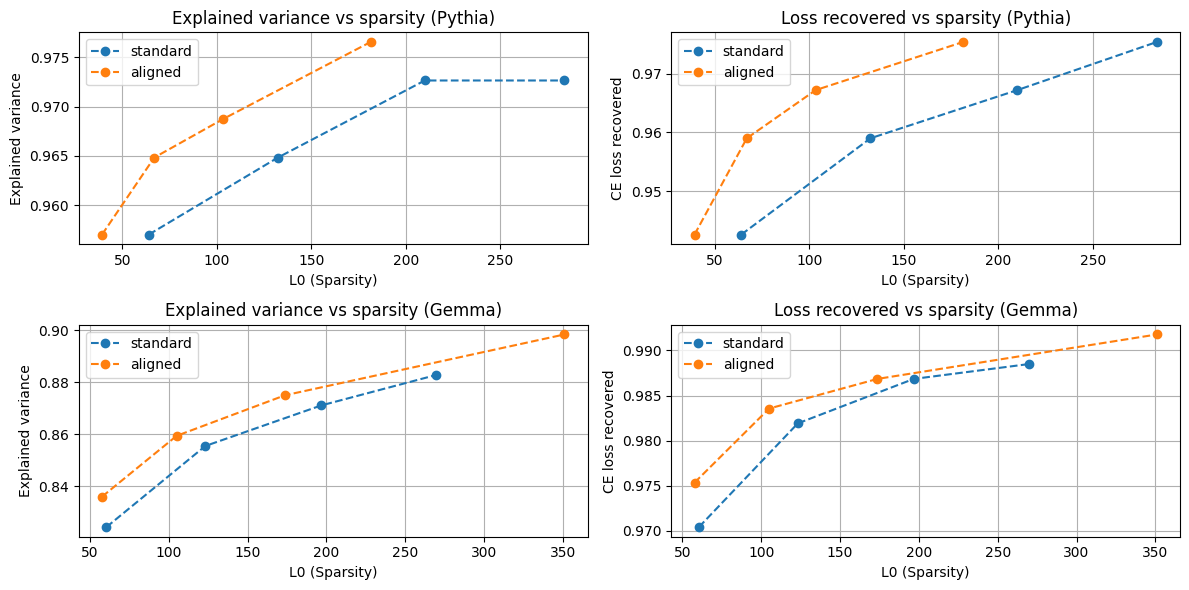}
    \caption{Reconstruction metrics, dictionary size 65K.}
    \label{fig:reconstruction 65 K}
\end{figure}

\begin{figure}[htb!]
    \centering
    \includegraphics[width=1.0\linewidth]{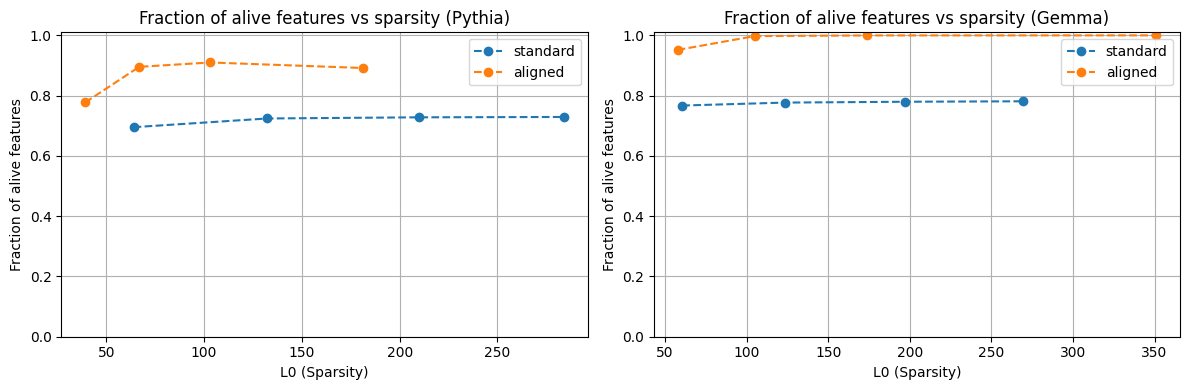}
    \caption{Alive-feature fraction, dictionary size 65K.}
    \label{fig:dead neurons 65 K}
\end{figure}

\section{Scaling to 500M Tokens and State-of-the-Art Comparison} \label{apx: 500M}

\begin{figure}[htb!]
    \centering
    \includegraphics[width=1.0\linewidth]{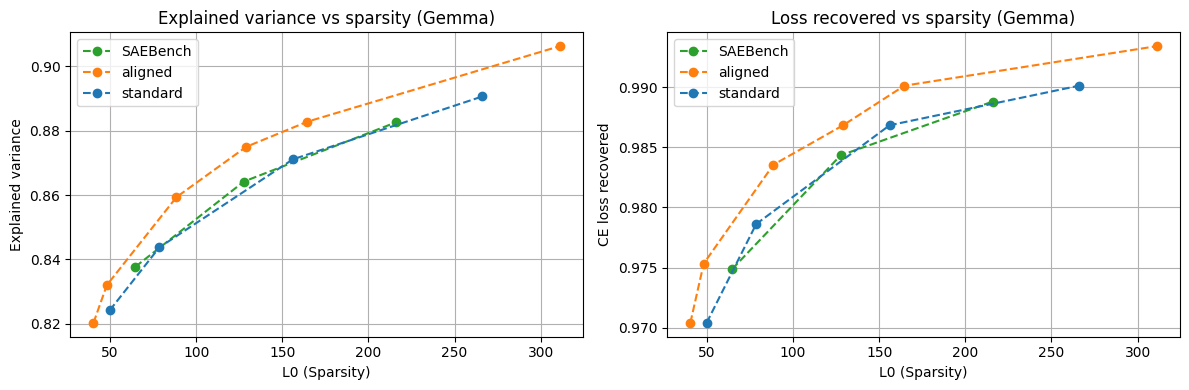}
    \caption{Reconstruction metrics at 500M tokens, dictionary size 65K, Gemma 2 2B (layer 12). Aligned training outperforms both our standard baseline and SAEBench state-of-the-art checkpoints.}
    \label{fig:reconstruction 65K 500M}
\end{figure}

\begin{figure}[htb!]
    \centering
    \includegraphics[width=0.8\linewidth]{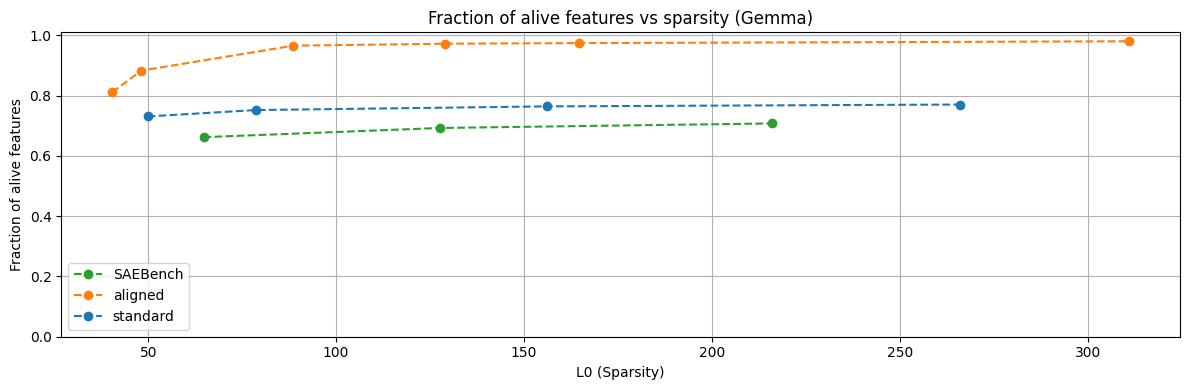}
    \caption{Alive-feature fraction at 500M tokens, dictionary size 65K, Gemma 2 2B (layer 12).}
    \label{fig:dead neurons 65K 500M}
\end{figure}

\end{document}